\documentclass[runningheads]{llncs}

\usepackage{iciap}

\usepackage{iciapabbrv}

\usepackage{graphicx}
\usepackage{booktabs}

\usepackage[accsupp]{axessibility}  

\usepackage{hyperref}

\usepackage{orcidlink}

\usepackage{amssymb}
\usepackage{amsmath}
\usepackage{pgfplots}
\pgfplotsset{compat=1.18}
\usepackage{booktabs}
\usepackage{multirow}
\usepackage{adjustbox}

\usepackage[dvipsnames]{xcolor}
\usepackage{xcolor-material}
\usepackage{color, colortbl}
\usepackage{soul}

\newenvironment{credits}{%
\begingroup\small%
\renewcommand\subsubsection{\@startsection{subsubsection}{3}{\z@}%
        {-12\p@ \@plus -4\p@ \@minus -4\p@}%
        {-0.5em \@plus -0.22em \@minus -0.1em}%
        {\normalfont\small\bfseries\boldmath}}
\renewcommand\paragraph{\@startsection{paragraph}{4}{\z@}%
        {-8\p@ \@plus -4\p@ \@minus -4\p@}%
        {-0.5em \@plus -0.22em \@minus -0.1em}%
        {\normalfont\small\itshape}}%
}{\endgroup}

\begin{document}
\title{FS-SAM2: Adapting Segment Anything Model~2 for Few-Shot Semantic Segmentation via Low-Rank Adaptation}
\titlerunning{FS-SAM2}

\author{
Bernardo Forni\inst{1}\orcidlink{0009-0007-3569-3432} \and 
Gabriele Lombardi\inst{2} \and
Federico Pozzi\inst{2} \and
Mirco Planamente\inst{2}\orcidlink{0000-0001-7238-1867}
}
\authorrunning{B.~Forni, G.~Lombardi, F.~Pozzi, M.~Planamente}
\institute{
University of Pavia, Pavia, Italy\\
\email{bernardo.forni01@universitadipavia.it}
\and
ARGO Vision, Milano, Italy\\
\email{\{gabriele.lombardi,federico.pozzi,mirco.planamente\}@argo.vision}
}
\maketitle  

\begin{abstract}

Few-shot semantic segmentation has recently attracted great attention.
The goal is to develop a model capable of segmenting unseen classes using only a few annotated samples.
Most existing approaches adapt a pre-trained model by training from scratch an additional module.
Achieving optimal performance with these approaches requires extensive training on large-scale datasets.
The Segment Anything Model 2 (SAM2) is a foundational model for zero-shot image and video segmentation with a modular design.
In this paper, we propose a \textbf{F}ew-\textbf{S}hot segmentation method based on \textbf{SAM2} (FS-SAM2), where SAM2's video capabilities are directly repurposed for the few-shot task.
Moreover, we apply a Low-Rank Adaptation (LoRA) to the original modules in order to handle the diverse images typically found in standard datasets, unlike the temporally connected frames used in SAM2's pre-training.
With this approach, only a small number of parameters is meta-trained, which effectively adapts SAM2 while benefiting from its impressive segmentation performance.
Our method supports any $K$-shot configuration.
We evaluate FS-SAM2 on the PASCAL-5$^i$, COCO-20$^i$ and FSS-$1000$ datasets, achieving remarkable results and demonstrating excellent computational efficiency during inference.
\\
Code is available at \href{https://github.com/fornib/FS-SAM2}{https://github.com/fornib/FS-SAM2}.


\end{abstract}

\section{Introduction}  \label{sec:intro}

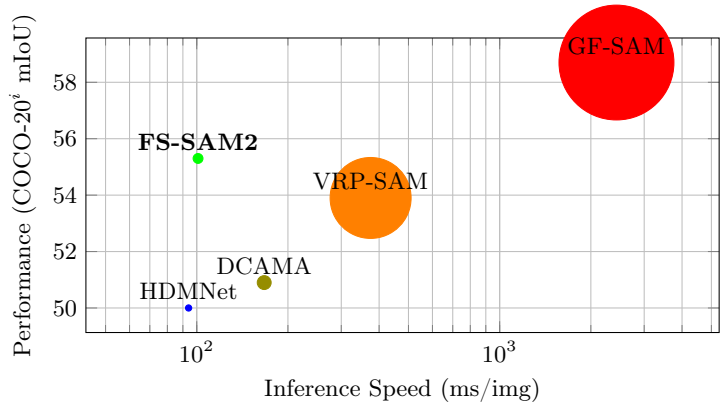
\begin{figure}[t]
\centering
\begin{tikzpicture}
  \begin{axis}[
    width=10cm, height=5.5cm,
    xlabel={Inference Speed (ms/img)},  
    ylabel={Performance (COCO-20$^i$ mIoU)},  
    xmode=log,
    grid=both,
    enlarge x limits=0.24,
    scatter/classes={
      FS-SAM2={mark=*,draw=green,fill=green},
      VRP-SAM={mark=*,draw=orange,fill=orange},
      DCAMA={mark=*,draw=olive,fill=olive},
      HDMNet={mark=*,draw=blue,fill=blue},
      Matcher={mark=*,draw=red,fill=red},
      GF-SAM={mark=*,draw=red,fill=red}
    },
  ]
  
  \addplot[
    scatter, only marks,
    scatter src=explicit symbolic,
    visualization depends on={value \thisrow{param} \as \perpointmarksize},
    scatter/@pre marker code/.append style={/tikz/mark size=0.023*\perpointmarksize},
    nodes near coords*={
      \ifnum\pdfstrcmp{\pgfplotspointmeta}{FS-SAM2}=0 \textbf{FS-SAM2} \else \pgfplotspointmeta \fi
    },
  ]
  table[meta=class] {
    time    performance   param  class
    101     55.3          81     FS-SAM2
    375     53.9          666    VRP-SAM
    167     50.9          114    DCAMA
    94      50.0          51     HDMNet
    2430    58.7          945    GF-SAM
  };
  \end{axis}
\end{tikzpicture}
\caption{
Performance-efficiency comparison of FSS models. Size is proportional to the number of parameters.}
\label{fig:bubble}
\end{figure}

\textit{Semantic Segmentation} is one of the most important tasks in Computer Vision.
The goal is to assign a predefined class label to each pixel in an image.
Recent developments in Deep Neural Networks have made tremendous progress in the semantic segmentation task \cite{csurka2022semantic}.
However, most advancements can be attributed to the availability of large amounts of training samples for each target class.
In real-world applications, providing labeled data for novel classes can be expensive or even impractical, especially when pixel-wise labeling is required.
In low-data scenarios, most methods exhibit poor generalization performance due to overfitting.
To deal with this challenging setting, \textit{Few-shot} Semantic Segmentation (FSS) aims to train a semantic segmentation model that adapts to novel classes using a limited amount of samples \cite{FSS}.
Specifically, the model is evaluated on segmentation performances for a novel class within a query image, given a support set that contains only a few annotated images of that class.
Recent approaches are based on a pre-trained image encoder to extract query and support features, which are then combined to decode the target segmentation \cite{HSNet,DCAMA,HDMNet,VRP-SAM,GF-SAM}.
These methods either train a complex and ad-hoc module with a relatively small number of training samples, or employ huge encoders to extract more general features.
The latter approach typically demonstrates better performances, but these performances are obtained at the cost of a significantly lower inference speed.
In practical applications, real-time latency is often a critical constraint, and it forces a trade-off between efficiency and performance.

The Segment Anything Model 2 (SAM2) \cite{SAM2} is a \textit{Vision Foundation Model} for class-agnostic image and video segmentation, that can utilize different kinds of user-provided prompts, such as points, boxes or masks.
Thanks to a powerful model architecture, based on a Hiera image encoder \cite{ryali2023hiera}, and exploiting a large-scale training process for video segmentation, SAM2 boasts strong performance with fast inference speed.
An internal module, called memory attention, enables SAM2 to track a prompted object across video frames by matching their features.
In this paper, we propose to leverage this matching capability for the few-shot task by treating support images as annotated video frames, and using the query image as the subsequent frame to be segmented.
We adapt SAM2's memory and encoder modules by employing Low-Rank Adaptation (LoRA), a fine-tuning strategy that introduces a small number of trainable parameters in each module.
This approach enables effective model adaptation and enhanced feature extraction without the need to train from scratch any ad-hoc module.

Our contributions are as follows:

\begin{enumerate}

    \item We propose FS-SAM2: a simple yet effective repurposing of the SAM2 architecture, demonstrating how its video-based design can be seamlessly adapted to the few-shot semantic segmentation task.
    
    \item We apply Low-Rank Adaptation (LoRA) to adapt SAM2 for efficient and robust learning in the few-shot setting with a simple training strategy and minimal parameter requirements.
            
    \item We conduct extensive empirical studies on PASCAL-$5^i$, COCO-$20^i$ and FSS-$1000$ datasets to validate our approach. FS-SAM2 achieves performance comparable to state-of-the-art methods while being computationally efficient.
    
\end{enumerate}

\section{Background and Related Works}

\subsection{Few-shot Semantic Segmentation}

FSS is the application of the few-shot learning paradigm \cite{fei2006one} to the semantic segmentation task, with the aim of improving generalization capabilities in low-data scenarios.
A \textit{support} set, consisting of few images with segmentation masks of a novel target class, is provided as input for model adaptation, and a \textit{query} image containing that class is used for evaluation.
Recent methods condition the query image on the novel class by matching query features with support features during inference.
In \textit{prototypical matching}, the target class is represented by one feature vector (or a few) extracted from the support set, which is used to classify each query feature \cite{PANet,PFENet,FPTrans}.
This heavy compression of the support set features helps with regularization, but may contain an incomplete or inaccurate representation of the target class.
Instead, \textit{pixel-wise matching} involves computing element-wise correlations between query features and support set features, thus providing more detailed context \cite{HSNet,CyCTR,VAT,DCAMA,SCCAN,HDMNet}.
In DCAMA \cite{DCAMA}, cross-attention is used to compute pixel-wise matching and aggregate support masks into multi-scale query features, which are fed into a convolutional decoder to extract the segmentation mask.

\subsection{SAM-based Few-Shot Segmentation}

SAM \cite{SAM} is a segmentation model composed of three modules: an image encoder that extracts embeddings from the input image, a prompt encoder that computes embeddings based on user-provided prompts, and a mask decoder that combines these embeddings to generate the segmentation mask.
SAM2 \cite{SAM2} extends the segmentation task to videos by adding a memory encoder module. It encodes image embeddings of previous frames with their predicted masks. A memory attention module conditions each new frame on these encoded embeddings using pixel-wise matching.
Recent methods \cite{VRP-SAM,Matcher,GF-SAM} successfully adapt SAM to the FSS setting by leveraging the prompt capabilities, although they often require an additional encoder to achieve satisfactory segmentation performance.
For example, VRP-SAM \cite{VRP-SAM} directly generates prompt embeddings that are applied to the query image by training a separate network based on PFENet \cite{PFENet} and the cross-attention of learnable vectors.
Similarly, GF-SAM \cite{GF-SAM} matches image features to generate a prior query mask, which is then refined by clustering SAM's output masks derived using points from the prior mask.
GF-SAM displays outstanding performance, but is overall very resource-intensive as it employs a separate DINOv2 encoder \cite{DINOv2} and repeatedly applies SAM's mask decoder.
Furthermore, GF-SAM does not perform any in-domain training but relies solely on the generalization capabilities of DINOv2, which has been shown \cite{bensaid2024novel} to outperform most semantic models.
Other methods \cite{MedSAM-2,RevSAM2} use SAM2 for medical images by repurposing the memory attention module with a selection criterion to identify a small subset of samples to serve as annotated frames.
These methods work without additional training, as medical datasets generally consist of similar images, analogous to video frames from SAM2's pre-training.

\subsection{Fine-tuning strategies}

Vision foundation models are pre-trained on large-scale datasets and excel at extracting high-quality image representations.
Most FSS methods use these models as the \textit{backbone} of the network, typically kept frozen to reduce overfitting and followed by an ad-hoc module, meta-trained for the few-shot task.
However, as shown in \cite{sun2022singular}, fine-tuning just a small amount of parameters in the backbone may improve model generalization on learning novel classes.
One of the most effective and widely adopted parameter-efficient fine-tuning methods is Low-Rank Adaptation (LoRA) \cite{LoRA,zanella2024low}.
This method consists of adapting selected linear layers by projecting initial features into a low-rank representation and back to output size for point-wise summation with the original output features.
\begin{equation}
\hat{y} = \hat{W}x = Wx + BAx
\end{equation}
where $\hat{y} \in \mathbb{R}^{d_{\text{out}}}$ is the updated layer output, $\mathbf{x} \in \mathbb{R}^{d_{\text{in}}}$ is the layer input, $W \in \mathbb{R}^{d_\text{out} \times d_{\text{in}}}$ is the weight matrix associated with the linear layer, $A \in \mathbb{R}^{r \times d_\text{in}}$ is the down-projection, $\mathbf{B} \in \mathbb{R}^{d_\text{out} \times r}$ is the up-projection, and $r \ll \min\{d_\text{in}, d_\text{out}\}$ is the rank.
LoRA minimizes the number of trainable parameters by fine-tuning only the low-rank projections while keeping the original parameters frozen.
After fine-tuning, the low-rank updates can be incorporated into existing layers by replacing the original weights $W$ with the updated ones $\hat{W}$.
This integration allows the model to utilize the benefits of fine-tuning without retaining separate LoRA components during inference, thereby maintaining the original model's computational efficiency and simplifying deployment.

\section{Methodology}

\subsection{Problem Setting}

FSS in the $K$-shot regime is the task of learning a segmentation model $\mathcal{F}_\Theta$ trained on a meta-training dataset $\mathcal{D}_{train}$ and evaluated on a separate meta-test dataset $\mathcal{D}_{test}$.
These datasets consist of images with associated segmentation masks of a single class, where the segmented classes are disjoint between the two datasets.
Following previous works \cite{fei2006one,PFENet,PANet}, the episodic learning paradigm is applied to both datasets $\mathcal{D}_{train}$ and $\mathcal{D}_{test}$, ensuring that the training process mimics the evaluation process.

For each dataset, let $l$ be a single class that is not the background.
The support set is defined as $S = \{(I_i, M_i)\}^K_{i=1}$, where $K$ is the number of shots indexed by $i$, $I_i \in \mathbb{R}^{H_i \times W_i \times 3}$ are different RGB images in the dataset containing class $l$ and $M_i \in \{0,1\}^{H_i \times W_i}$ are their corresponding binary segmentation mask for class $l$.
The query image $I_q \in \mathbb{R}^{H_q \times W_q \times 3}$ is an RGB image from the same dataset but distinct from those in $S$, and the query mask $M_q \in \{0,1\}^{H_q \times W_q}$ is the corresponding mask for class $l$.
An episode consists of a support set together with a query image.
For each episode, the task is to generalize from the support set and accurately segment the class $l$ in the query image.
\begin{equation}
    \hat{M_q} = \mathcal{F}_\Theta(I_q, S) \in \{0,1\}^{H_q \times W_q}
    \quad\quad\quad\quad
    \text{IoU}_l =  \frac{\sum_{q} |M_q \cap \hat{M_q}|}{\sum_{q} |M_q \cup \hat{M_q}|}
\end{equation}
where $\hat{M_q}$ is the predicted binary segmentation of the query image for class $l$.
During evaluation on the meta-test dataset, the episodes are independent of each other, ensuring no information is shared between them.
The standard evaluation metric for FSS is mean Intersection-over-Union (mIoU):
the $\text{IoU}$ is calculated as a pixel sum over all query images $q$, the mIoU is computed as the average IoU across all foreground classes: $\text{mIoU} = \frac{1}{C} \sum_{l=1}^{C} \text{IoU}_l $ where $C$ is the total number of classes in the evaluation dataset.

\subsection{FS-SAM2}  
In this work, we present a simple yet effective strategy to adapt the SAM2 for the FSS task, taking advantage of its extensive pre-training and computational efficiency.
We repurpose SAM2's video segmentation capabilities by treating the support set $S$ as annotated video frames and the query image $I_q$ as the subsequent frame to be segmented.
We notice that SAM2's pre-training, based on temporally connected frames that are very similar to each other, heavily relies on these frame similarities.
This results in suboptimal FSS performance on standard datasets, where the support and query images can significantly differ.
To address this issue, we employ a meta-training strategy to learn a small amount of parameters and effectively adapt the model.

\begin{figure}[t]
    \centering
    \includegraphics[width=\linewidth]{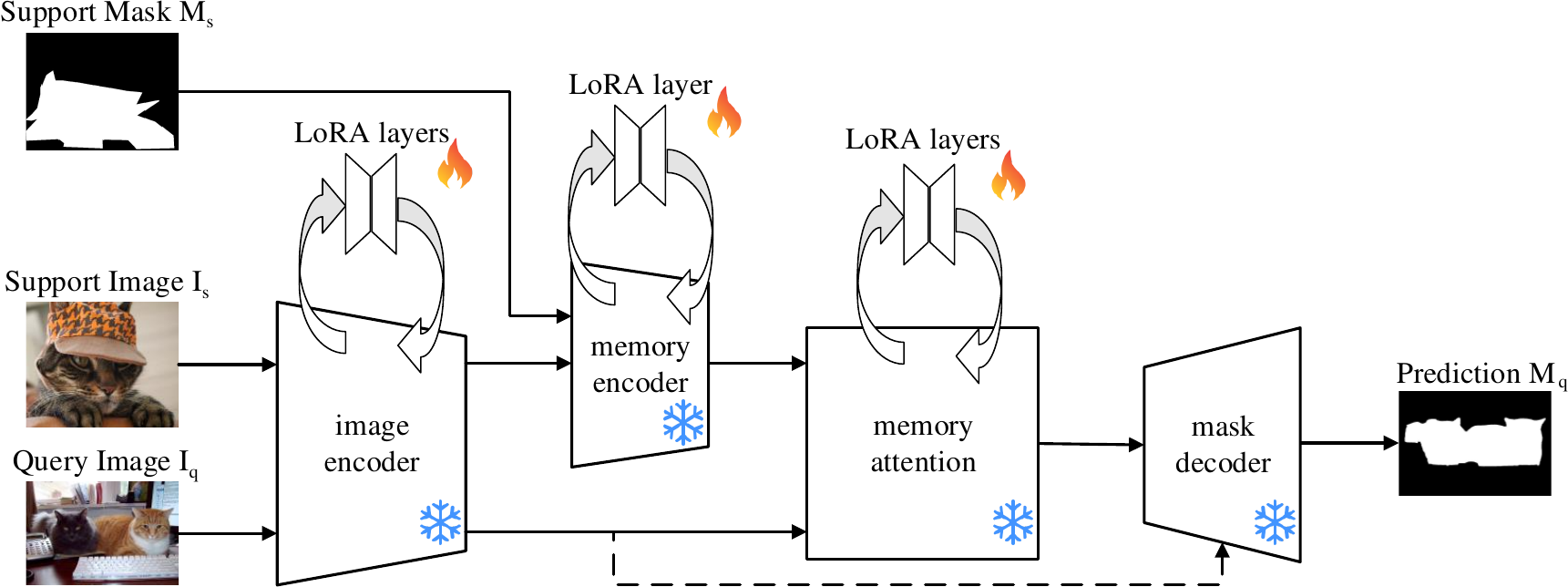}
    \caption{
    FS-SAM2 framework for 1-shot semantic segmentation.
    The SAM2 video segmentation pipeline is repurposed by treating the support and query images as consecutive input frames, taking advantage of pixel-wise matching in the \textit{memory attention} module.
    The model is efficiently specialized by meta-training only LoRA layers.
    }
    \label{fig:framework}
\end{figure}

The overall architecture of FS-SAM2 is illustrated in Fig. \ref{fig:framework}, the modules are all inherited from SAM2.
Given a support image $I_s$ with its corresponding mask $M_s$ and a query image $I_q$, high-quality features are independently extracted from both images using the \textit{image encoder}.
The support mask and features are downsampled and combined within the \textit{memory encoder}, a convolutional module that incorporates the segmentation mask into the image features.
The \textit{memory attention} conditions the query features on the encoded support features through pixel-wise matching, utilizing repeated cross-attention and self-attention operations on the image embedding.
The output embedding with high-level query features is used to generate the prediction mask $M_q$ in the \textit{mask decoder}, a module based on a lightweight transformer architecture \cite{dosovitskiy2020image} with convolutional upsampling.
The prompt encoder is ignored, as only default token embeddings are used for the mask decoder and geometric prompts are not required.

We apply LoRA to image encoder, memory encoder and memory attention.
The mask decoder is kept as-is, since its extensive class-agnostic pre-training makes it already suitable for the task.
During meta-training, the unlabeled query image is matched with labeled support images using the memory attention module.
The predicted labels for the query image are then compared to the ground-truth labels to compute the loss, which is used to fine-tune the parameters.
This approach allows fine-tuning by meta-training only a small fraction of parameters, while effectively specializing the model for the few-shot setting.

\subsection{Extension to $K$-shots}

Our approach accommodates any $K$-shot configuration without the need to train a different model for each $K$.
We treat all the $K$ support images as previously annotated frames and apply the same framework.
Specifically, each support image and mask is encoded by means of the \textit{memory encoder}, these encodings are then concatenated along the spatial dimension before applying the \textit{memory attention}.

\section{Experimental Results}  

\subsection{Datasets}

We evaluate FS-SAM2 performance and generalization capabilities by conducting extensive experiments on three widely used public datasets: PASCAL-$5^i$ \cite{FSS}, with $20$ classes from the PASCAL VOC 2012 \cite{PASCAL} and Extended SDS \cite{hariharan2011semantic} datasets; COCO-$20^i$ \cite{FWB,PANet}, which is based on the MS COCO \cite{COCO} dataset with $80$ classes; and FSS-$1000$ \cite{FSS1000}, which contains $1000$ classes.
PASCAL-$5^i$ and COCO-$20^i$ classes are evenly split in $4$ folds evaluated using cross-validation, while FSS-$1000$ is split into $520$ classes for training, $240$ for validation,
and $240$ for testing.
Following recent works \cite{PFENet,HSNet}, we also evaluate the domain-shift scenario, where each model trained on a COCO-$20^i$ fold is tested on PASCAL-VOC images that do not overlap in classes for that fold.
To ensure consistency with other benchmarks, we conducted tests using 1000 randomly sampled episodes for each fold.

To produce an episode, the query image is first selected, then a class in the image is randomly sampled, and finally a support set corresponding to that class is randomly sampled, where other classes are marked as background in the segmentation masks.
For COCO-$20^i$, following previous works \cite{HSNet,DCAMA,VRP-SAM,GF-SAM}, the class is first randomly sampled, and then a query image and a support set are randomly sampled with that class.

\subsection{Implementation details}

We mostly follow the implementation of VRP-SAM \cite{VRP-SAM}.
We use Binary Cross-Entropy (BCE) loss and Dice loss to supervise the learning during meta-training.
We employ the AdamW optimizer \cite{loshchilov2017decoupled} ($\beta_1=0.9,\beta_2=0.999$) with a cosine learning rate decay schedule.
The model is trained for $50$ epochs with an initial learning rate of $1e^{-4}$ and batch size of $16$.
Each epoch corresponds to the size of the meta-train dataset, with early stopping validation on the meta-test dataset.
We adopt the same training augmentations as some previous works \cite{PFENet,HDMNet}.
Images are first resized to $1024 \times 1024$ before being processed by SAM2, the predicted masks are resized to their original resolution for evaluation.
The performance of the model is evaluated using mean Intersection-over-Union (mIoU) for each fold.

We conducted all the experiments on the Hiera-B+ version of SAM2 initialized with SAM2.1 pre-trained weights.
We apply LoRA to the Q, V, K, O projection layers of the transformer blocks.
We set the LoRA rank to $4$ in the image encoder,
while in the memory attention and memory encoder we use a rank of $32$ to introduce a comparable amount of trainable parameters given the small amount of layers.
The architecture totals $81$ million parameters, with only $750$ thousand trainable.
The training was done with two NVIDIA A100-64GB GPUs and casting to bfloat16.

\subsection{Achieved Results}  

\begin{table}[t]
    \caption{Performance results for FSS on PASCAL-$5^i$ dataset with mIoU metric.
    The best results are highlighted in \textbf{bold} and the second-best results are \underline{underlined}.
    }
    \label{table:pascal}
    \centering
    \resizebox{\textwidth}{!}{%
    \begin{tabular}{ll ccccc | ccccc}
        \toprule
        \multirow{2}{*}{Method} & \multicolumn{1}{l}{\multirow{2}{*}{Backbone}} & \multicolumn{5}{c}{PASCAL-$5^i$ 1-shot} & \multicolumn{5}{c}{PASCAL-$5^i$ 5-shot} \\
        \cmidrule(lr){3-7} \cmidrule(lr){8-12}
        & & fold-0 & fold-1 & fold-2 & fold-3 & \multicolumn{1}{c}{mean} & fold-0 & fold-1 & fold-2 & fold-3 & \multicolumn{1}{c}{mean} \\
        \midrule
        PFENet\cite{PFENet} & RN50 & 61.7 & 69.5 & 55.4 & 56.3 & 60.8 & 63.1 & 70.7 & 55.8 & 57.9 & 61.9 \\
        HSNet\cite{HSNet} & RN50 & 64.3 & 70.7 & 60.3 & 60.5 & 64.0 & 70.3 & 73.2 & 67.4 & 67.1 & 69.5 \\
        DCAMA\cite{DCAMA} & Swin & 72.2 & 73.8 & 64.3 & \underline{67.1} & 69.3 & 75.7 & 77.1 & 72.0 & 74.8 & 74.9 \\
        FPTrans\cite{FPTrans} & DeiT & 72.3 & 70.6 & 68.3 & 64.1 & 68.8 & \underline{76.7} & 79.0 & \textbf{81.0} & \underline{75.1} & \underline{78.0} \\
        Matcher\cite{Matcher} & SAM,DINOv2 & 67.7 & 70.7 & 66.9 & 67.0 & 68.1 & 71.4 & 77.5 & 74.1 & 72.8 & 74.0 \\
        HDMNet\cite{HDMNet} & RN50 & 71.0 & 75.4 & 68.9 & 62.1 & 69.4 & 71.3 & 76.2 & 71.3 & 68.5 & 71.8 \\
        VRP-SAM\cite{VRP-SAM} & SAM,RN50 & \underline{73.9} & \underline{78.3} & \underline{70.6} & 65.0 & 71.9 & - & - & - & - & 72.9 \\
        GF-SAM\cite{GF-SAM} & SAM,DINOv2 & 71.1 & 75.7 & 69.2 & \textbf{73.3} & \underline{72.1} & \textbf{81.5} & \textbf{86.3} & \underline{79.7} & \textbf{82.9} & \textbf{82.6} \\
        \midrule
        VRP-SAM2 & SAM2,RN50 & 74.0 & 77.4 & 69.3 & 63.5 & 71.0 & 75.6 & 74.6 & 69.7 & 61.8 & 70.4 \\
        \textbf{FS-SAM2} & SAM2 & \textbf{74.0} & \textbf{79.6} & \textbf{73.5} & 66.7 & \textbf{73.4} & 75.7 & \underline{79.6} & 75.3 & 68.8 & 74.8 \\

        \bottomrule
    \end{tabular}%
    }
\end{table}

\begin{table}[t]
    \caption{Performance results for FSS on the COCO-$20^i$ dataset with mIoU metric.
    The best results are highlighted in \textbf{bold} and the second-best results are \underline{underlined}.
    }
    \label{table:coco}
    \centering
    \resizebox{\textwidth}{!}{%
    \begin{tabular}{ll ccccc | ccccc}
        \toprule
        \multirow{2}{*}{Method} & \multicolumn{1}{l}{\multirow{2}{*}{Backbone}} & \multicolumn{5}{c}{COCO-$20^i$ 1-shot} & \multicolumn{5}{c}{COCO-$20^i$ 5-shot} \\
        \cmidrule(lr){3-7} \cmidrule(lr){8-12}
        & & fold-0 & fold-1 & fold-2 & fold-3 & \multicolumn{1}{c}{mean} & fold-0 & fold-1 & fold-2 & fold-3 & mean \\
        \midrule
        PFENet\cite{PFENet} & RN50 & 36.5 & 38.6 & 34.5 & 33.8 & 35.8 & 36.5 & 43.3 & 37.8 & 38.4 & 39.0 \\
        HSNet\cite{HSNet} & RN50 & 36.3 & 43.1 & 38.7 & 38.7 & 39.2 & 43.3 & 51.3 & 48.2 & 45.0 & 46.9 \\
        DCAMA\cite{DCAMA} & Swin & 49.5 & 52.7 & 52.8 & 48.7 & 50.9 & 55.4 & 60.3 & 59.9 & 57.5 & 58.3 \\
        FPTrans\cite{FPTrans} & DeiT & 44.4 & 48.9 & 50.6 & 44.0 & 47.0 & 54.2 & 62.5 & \underline{61.3} & 57.6 & 58.9 \\
        Matcher\cite{Matcher} & SAM,DINOv2 & \underline{52.7} & 53.5 & 52.6 & 52.1 & 52.7 & \underline{60.1} & \underline{62.7} & 60.9 & 59.2 & \underline{60.7} \\
        HDMNet\cite{HDMNet} & RN50 & 43.8 & 55.3 & 51.6 & 49.4 & 50.0 & 50.6 & 61.6 & 55.7 & 56.0 & 56.0 \\
        VRP-SAM\cite{VRP-SAM} & SAM,RN50 & 48.1 & 55.8 & \textbf{60.0} & 51.6 & 53.9 & - & - & - & - & - \\
        GF-SAM\cite{GF-SAM} & SAM,DINOv2 & \textbf{56.6} & \textbf{61.4} & \underline{59.6} & \textbf{57.1} & \textbf{58.7} & \textbf{67.1} & \textbf{69.4} & \textbf{66.0} & \textbf{64.8} & \textbf{66.8} \\
        \midrule
        VRP-SAM2 & SAM2,RN50 & 47.1 & 55.5 & 55.2 & 52.8 & 52.7 & 50.8 & 60.0 & 57.7 & 57.7 & 56.6 \\
        \textbf{FS-SAM2} & SAM2 & 49.3 & \underline{57.6} & 58.3 & \underline{56.1} & \underline{55.3} & 55.1 & 60.6 & 58.7 & \underline{59.3} & 58.4 \\
        \bottomrule
    \end{tabular}%
    }
\end{table}

\begin{table}[t]
    \caption{Performance results for FSS on FSS-$1000$ dataset and domain-shift from COCO-$20^i$ to PASCAL-VOC dataset with mIoU metric.
    The best results are highlighted in \textbf{bold} and the second-best results are \underline{underlined}.
    }
    \label{table:shift}
    \centering
    \begin{tabular}{ll cc | cc}
        \toprule
        \multirow{2}{*}{Method} & \multicolumn{1}{l}{\multirow{2}{*}{Backbone}} & \multicolumn{2}{c}{FSS-1000} & \multicolumn{2}{c}{domain-shift} \\
        \cmidrule(lr){3-4} \cmidrule(lr){5-6}
        & & 1-shot & 5-shot & 1-shot & 5-shot \\
        \midrule
        PFENet\cite{PFENet} & RN50 & 80.8 & 81.4 & 61.1 & 63.4 \\
        HSNet\cite{HSNet} & RN50 & 85.5 & 87.8 & 61.6 & 68.7 \\
        VAT\cite{VAT} & RN50 & \underline{90.1} & \underline{90.7} & 64.5 & 69.7 \\
        DCAMA\cite{DCAMA} \ & Swin & 88.2 & 88.8 & - & - \\
        FPTrans\cite{FPTrans} \ & DeiT & - & - & 69.7 & 79.3 \\
        MSDNet \cite{MSDNet} & RN50 & - & - & \underline{72.1} & \underline{74.2} \\
        \midrule
        \textbf{FS-SAM2} & SAM2 & \textbf{90.6} & \textbf{91.7} & \textbf{77.2} & \textbf{79.7} \\
        \bottomrule
    \end{tabular}%
\end{table}

\begin{figure*}[t]%
    \centering
    \begin{tabular}{ccccc}
        \textbf{Support} & \textbf{GT} & \textbf{VRP-SAM} & \textbf{SAM2} & \textbf{FS-SAM2} \\
        
        \includegraphics[width=0.19\textwidth]{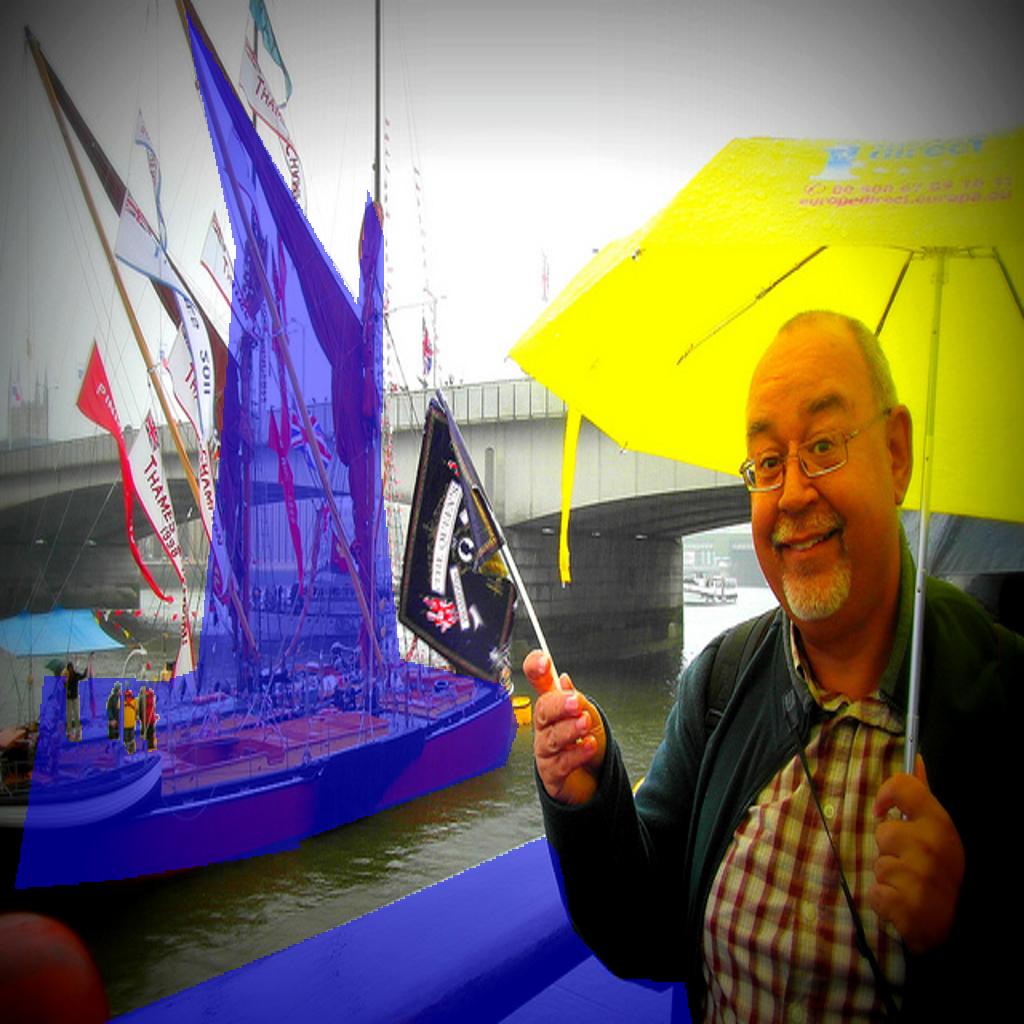} &
        \includegraphics[width=0.19\textwidth]{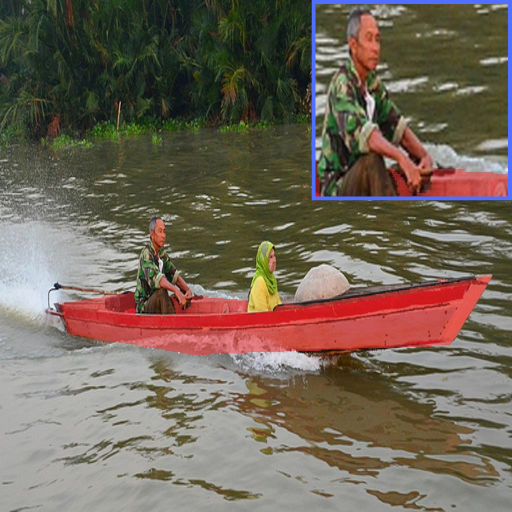} &
        \includegraphics[width=0.19\textwidth]{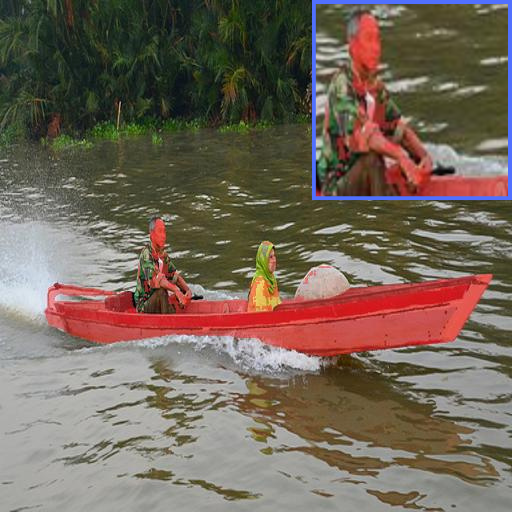} &
        \includegraphics[width=0.19\textwidth]{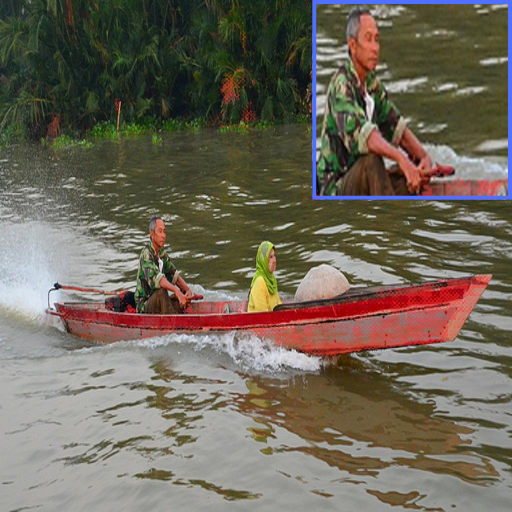} &
        \includegraphics[width=0.19\textwidth]{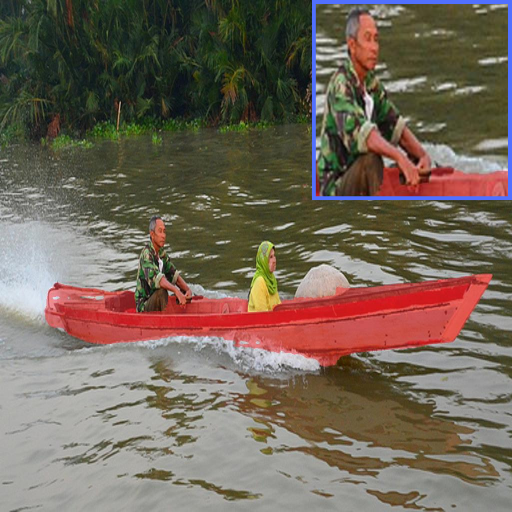} \\
        
        \includegraphics[width=0.19\textwidth]{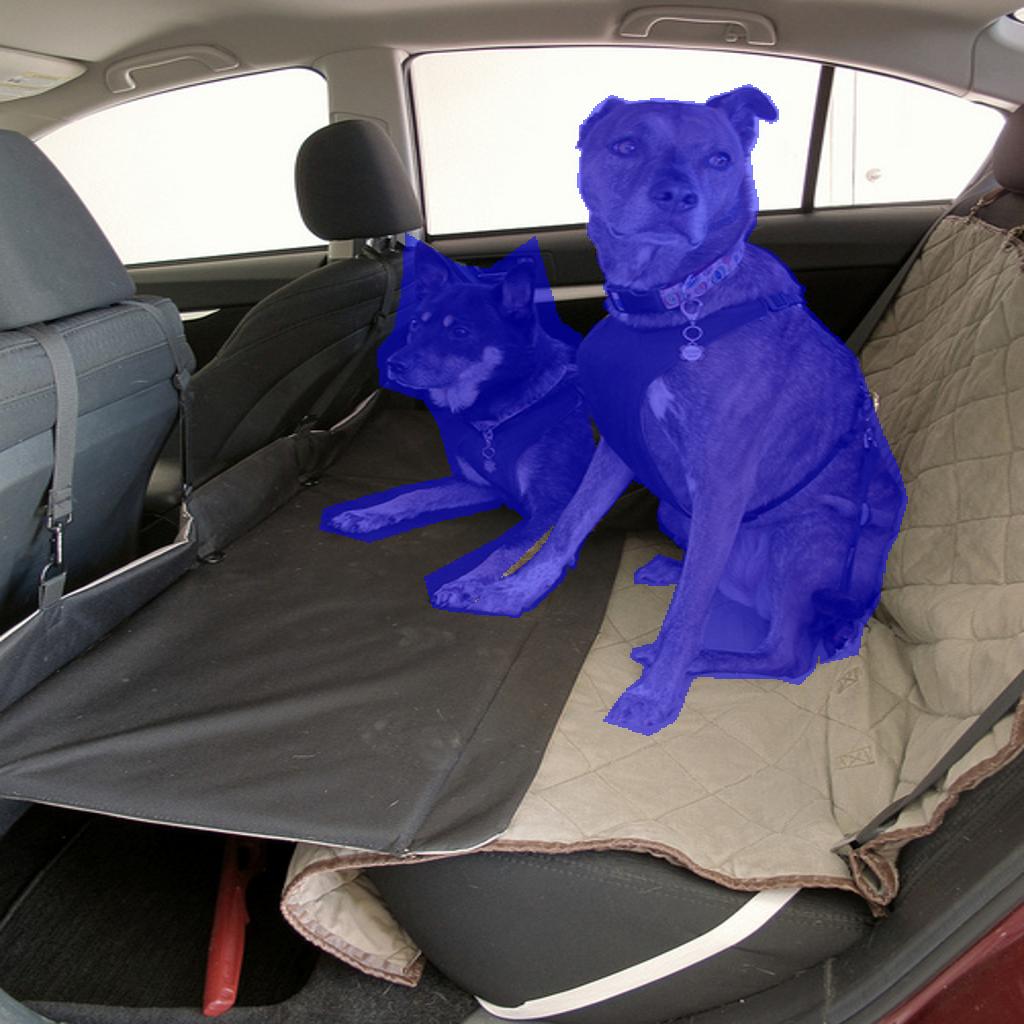} &
        \includegraphics[width=0.19\textwidth]{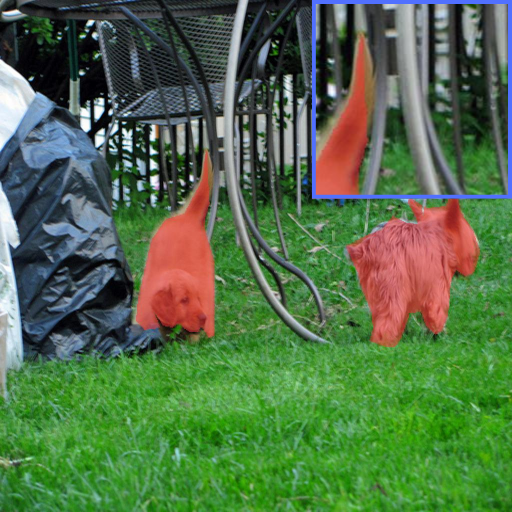} &
        \includegraphics[width=0.19\textwidth]{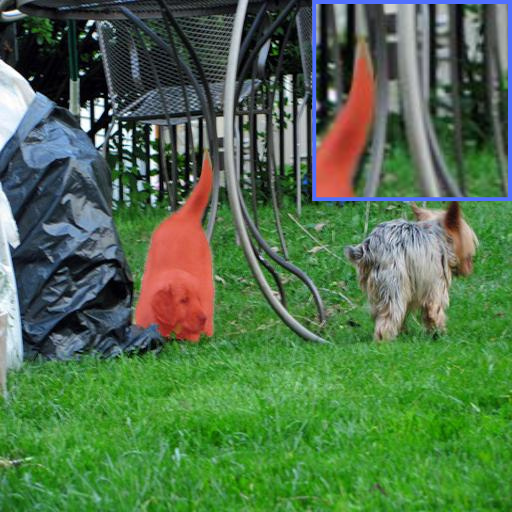} &
        \includegraphics[width=0.19\textwidth]{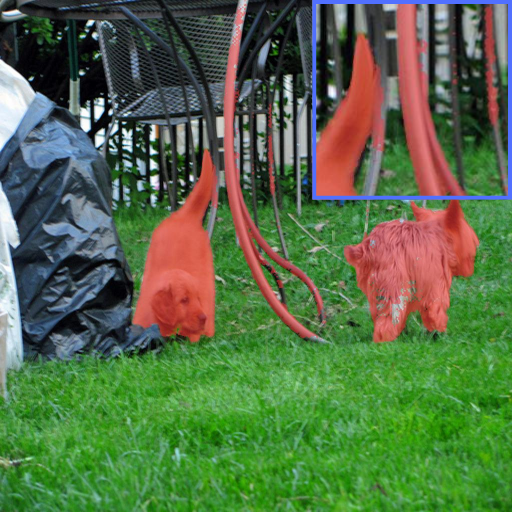} &
        \includegraphics[width=0.19\textwidth]{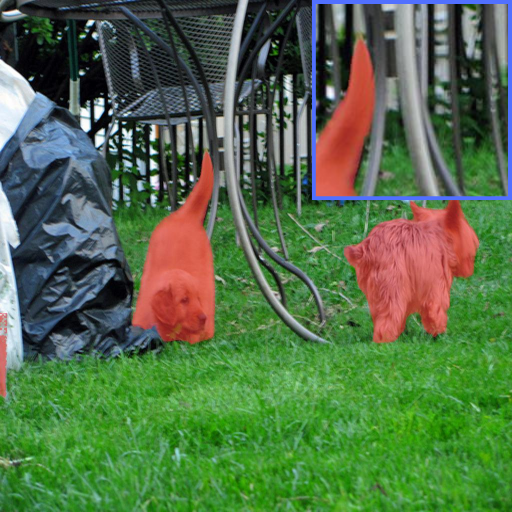} \\
    \end{tabular}
    \caption{Qualitative comparison results on the COCO-$20^i$ dataset.
    Our meta-training strategy successfully adapts SAM2, outperforming baseline methods.
    }
    \label{fig:qualitative}
\end{figure*}

We present performance comparisons of FS-SAM2 against other state-of-the-art (SOTA) methods for a comprehensive evaluation.
To further illustrate the improvements of our method, we also compare with VRP-SAM2, which is our implementation of VRP-SAM \cite{VRP-SAM} using SAM2 as the backbone instead of SAM.

Table \ref{table:pascal} presents performance comparisons on the PASCAL-$5^i$ dataset.
In the $1$-shot scenario, our model achieves a new SOTA mIoU of $73.4\%$, surpassing VRP-SAM by $1.5\%$, the most comparable method to ours.
The VRP-SAM method does not benefit from the newer SAM2 backbone, as it likely requires some fine-tuning of the backbone for the few-shot task.
It is worth noting that GF-SAM \cite{GF-SAM}, which utilizes a DINOv2-L \cite{DINOv2} backbone pre-trained on a dataset containing PASCAL-VOC, still underperforms compared to our method in the $1$-shot setting.
We observe a slight improvement in segmentation performance in the $5$-shot setting, indicating that our method can leverage contextual information from multiple support images.
However, further gains could be achieved by explicitly training the model in a $5$-shot regime.

Table \ref{table:coco} shows results on the COCO-$20^i$ dataset and demonstrates the effectiveness and generalization capabilities of FS-SAM2.
Our model outperforms VRP-SAM \cite{VRP-SAM} by $1.4\%$ in the $1$-shot benchmark and shows a notable increase in the $5$-shot benchmark.
While GF-SAM \cite{GF-SAM} is the current SOTA on this benchmark, we note that it does not undergo meta-training for in-domain knowledge but instead relies on DINOv2 pre-training and is significantly more resource-intensive than FS-SAM2, as illustrated in Fig. \ref{fig:bubble}.

We also evaluate our method on the FSS-$1000$ dataset, that contains a large number of different classes, and on the domain-shift scenario, where the base training data and the testing data have a significant domain gap.
The results presented in Table \ref{table:shift} clearly show the robustness of FS-SAM2 and demonstrate that LoRA successfully adapted the model to the few-shot setting.

Figure \ref{fig:qualitative} presents a qualitative comparison of our approach with VRP-SAM and the SAM2 without meta-training.
The examples highlight how the SAM2 struggles with dissimilar support-query pairs.
In contrast, FS-SAM2 consistently generates accurate and complete masks, capturing fine details and entire object regions that VRP-SAM misses.
Our method often generates cleaner boundaries, even refining minor inaccuracies present in the ground truth (GT) annotations.

\subsection{Ablation Study}

\begin{table}[t]
\begin{minipage}[t]{0.61\textwidth}

\caption{Performance comparison of different fine-tuning strategies on COCO-$20^i$ $1$-shot task averaged over 4 folds.
The LoRA rank applied to the module is indicated in parentheses.
The column \#params indicates the number of parameters updated during meta-training.}
\label{table:ft}
\centering
\begin{tabular}{l|c|r}
\hline
Strategy & \ mIoU \ & \ \#params \\ \hline
No training & 26.2 & 0 \\
Fine-tuning mem. & 51.8 & 7,306,912 \\
LoRA mem.(32) & 49.7 & 475,136 \\
LoRA enc.(4) & 54.6 & 265,356 \\
LoRA enc.(32) & 54.8 & 1,536,560 \\
LoRA enc.(4), mem.(4) & \underline{55.2} & 326,028 \\
LoRA enc.(4), mem.(32) & \textbf{55.3} & 750,732 \\
LoRA enc.(4), mem.(32), dec.(32) & 55.1 & 1,127,564 \\
LoRA enc.(4), fine-tuning mem. & 54.2 & 7,572,268 \\
\hline
\end{tabular}
\end{minipage}%
\hfill
\begin{minipage}[t]{0.36\textwidth}

\caption{Same as Table \ref{table:ft}, but the support image is set to be identical to the query image during the meta-test.}
\label{table:ft2}
\centering
\begin{tabular}{l|c}
\hline
Strategy & mIoU \\ \hline
No training & \textbf{95.4} \\
Fine-tuning mem. & 64.0 \\
LoRA mem.(32) & \underline{85.2} \\
LoRA enc.(4) & 82.5 \\
LoRA enc.(4), mem.(32) & 80.0 \\ \hline
\end{tabular}
\end{minipage}
\end{table}

We conducted ablation studies on the COCO-$20^i$ dataset in $1$-shot regime to analyze the contributions of various adaptation strategies.
This analysis aims to provide a deeper understanding of the role and importance of each module.

Table \ref{table:ft} compares different fine-tuning strategies applied to various modules of the model.
We observe that the model performs inadequately without meta-training.
Fine-tuning only the memory modules results in a significant improvement, demonstrating that the model can be effectively adapted for the FSS task.
Utilizing LoRA only in the image encoder yields improved performance, indicating that this strategy reduces the importance of feature position information for subsequent pixel-wise matching while improving generalization for novel classes.
Applying LoRA to both the image encoder and the memory modules results in the most effective strategy, achieving 55.3$\%$ mIoU.
Using a smaller rank of 4 for LoRA in the memory modules, or extending LoRA to the mask decoder, does not significantly impact performance.

Table \ref{table:ft2} compares fine-tuning strategies in the scenario where the support set is not randomly sampled but is the same as the query image and mask during the evaluation on the meta-test dataset.
This setup more closely mimics SAM2's original pre-training of similar frames and is beneficial for assessing the impact of meta-training.
Without any meta-training the performance is nearly perfect.
Moreover, we note that output masks are qualitatively better than the input masks.
Fine-tuning the memory modules severely deteriorates performance, indicating that the model has reduced the reliance on the similarity of the images when matching novel classes.
Applying LoRA better retains the learned capabilities while maintaining high performance on the FSS task.

\section{Conclusions}

In this paper, we introduced FS-SAM2, a new framework for SAM2 adaptation to the few-shot semantic segmentation task.
SAM2's video segmentation capabilities can be effectively repurposed to condition the query image on the support set by training a small number of parameters and without requiring additional backbones or ad-hoc modules.
Adaptation with LoRA further improves generalization performance on novel classes.
Overall, our method demonstrates very competitive results while being computationally efficient.

For future research directions, it would be worthwhile to investigate alternative $5$-shot training methods and explore more parameter-efficient fine-tuning strategies.
Furthermore, we aim to extend our approach to the multi-class setting, a relatively under-explored task that offers significant practical advantages, as SAM2 lacks inter-class communication capabilities.

\begin{credits}
\textbf{Acknowledgements.}
We acknowledge ISCRA for awarding this project access to the LEONARDO supercomputer, owned by the EuroHPC Joint Undertaking, hosted by CINECA (Italy).
\end{credits}

%
%
\bibliographystyle{splncs04}
\bibliography{main}

\end{document}